\documentclass[runningheads]{llncs}
\usepackage{graphicx}
\usepackage{amsmath}
\usepackage{amsfonts,amssymb}
\usepackage{subcaption}
\usepackage{etoolbox}

\newcommand{\repthanks}[1]{\textsuperscript{\ref{#1}}}

\makeatletter
\renewcommand*{\@fnsymbol}[1]{\ifcase#1\or*\else\dag \fi}
\makeatother

\makeatletter
\patchcmd{\maketitle}
  {\def\thanks}
  {\let\repthanks\repthanksunskip\def\thanks}
  {}{}
\patchcmd{\@maketitle}
  {\def\thanks}
  {\let\repthanks\@gobble\def\thanks}
  {}{}
\newcommand\repthanksunskip[1]{\unskip{}}
\makeatother

\newcommand{\argmin}{\operatornamewithlimits{argmin}}
\begin{document}
\title{Domain aware medical image classifier interpretation by counterfactual impact analysis\thanks{VRVis is funded by BMK, BMDW, Styria, SFG and Vienna Business Agency in the scope of COMET - Competence Centers for Excellent Technologies (854174) which is managed by FFG. Thanks go to our project partner AGFA HealthCare for providing valuable input.}}
\titlerunning{Domain aware classifier interpretation by counterfactual impact analysis}
%

\author{Dimitrios Lenis\and
David Major\and
Maria Wimmer\and
Astrid Berg\and
Gert Sluiter\and
Katja B\"uhler}
\institute{VRVis Zentrum f\"ur Virtual Reality und Visualisierung Forschungs-GmbH,\\Vienna, Austria}


\authorrunning{D. Lenis et al.}

\maketitle              
\begin{abstract}
The success of machine learning methods for computer vision tasks has driven a surge in computer assisted prediction for medicine and biology. Based on a data-driven relationship between input image and pathological classification, these predictors deliver unprecedented accuracy. Yet, the numerous approaches trying to explain the causality of this learned relationship have fallen short: time constraints, coarse, diffuse and at times misleading results, caused by the employment of heuristic techniques like Gaussian noise and blurring, have hindered their clinical adoption.

In this work, we discuss and overcome these obstacles by introducing a neural-network based attribution method, applicable to any trained predictor. Our solution identifies salient regions of an input image in a single forward-pass by measuring the effect of local image-perturbations on a predictor's score. We replace heuristic techniques with a strong neighborhood conditioned inpainting approach, avoiding anatomically implausible, hence adversarial artifacts. We evaluate on public mammography data and compare against existing state-of-the-art methods. Furthermore, we exemplify the approach's generalizability by demonstrating results on chest X-rays. Our solution shows, both quantitatively and qualitatively, a significant reduction of localization ambiguity and clearer conveying results, without sacrificing time efficiency.


\end{abstract}
\section{Introduction}
The last decade's success of machine learning methods for computer-vision tasks has driven a surge in computer assisted prediction for medicine and biology. This has posed a conundrum. Current predictors, predominantly artificial neural networks (ANNs), learn a data-driven relationship between input image and pathological classification, whose validity, i.e. accuracy and specificity, we can quantitatively test. In contrast, this learned relationship's causality typically remains elusive \cite{adadi2018,lipton2018,litjens2017}. A plethora of approaches have been proposed that aim to fill this gap by explaining causality through identifying and attributing salient image-regions responsible for a predictor's outcome \cite{chang2018,dabkowski2017,fong2019,simonyan2013,selvaraju2017,uzunova2019}.   

\begin{figure}[t]
\centering
\begin{subfigure}[b]{0.35\textwidth}
\includegraphics[width=\textwidth]{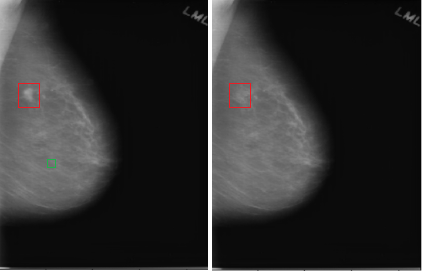}
\caption{}\label{fig:1a}
\end{subfigure}
\begin{subfigure}[b]{0.25\textwidth}
\centering
\includegraphics[width=0.9\textwidth]{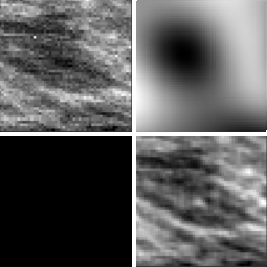}
\caption{}\label{fig:1b}
\end{subfigure}
\begin{subfigure}[b]{0.3\textwidth}
\centering
\includegraphics[width=0.9\textwidth,trim={10 35 0 0},clip]{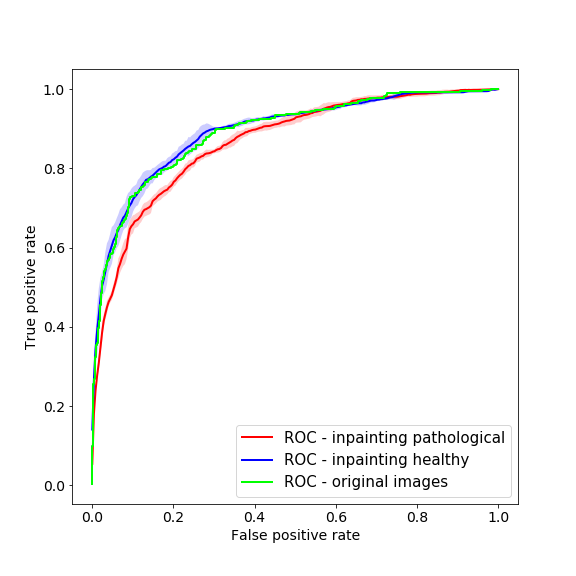}
\caption{}\label{fig:1b}
\end{subfigure}

\caption{Overview of marginalization: (a) original with annotated mass (red box) before and after marginalization by our method; (b) local comparisons with popular methods (clockwise): original, blurring~\cite{fong2019}, inpainting (ours), and averaging~\cite{zeiler2017}; (c) ROC curves of the mammography classifier (green curve) vs. healthy pixel inpainting only in healthy/pathological (blue/red curves) structures.}
\label{fig:inpainting_ov}
\end{figure}

Lacking a canonical mapping between an ANN's prediction and its domain, this form of reasoning is predominantly based on \emph{local explanations} (LE), i.e. explicit attribution-maps characterizing image-prediction tuples \cite{lipton2018,fong2019}. Typically, these maps are loosely defined as regions with \emph{maximal influence} towards the predictor, implying that any texture change within the attributed area will significantly change the prediction. Besides technical insight, these LE can provide a key benefit for clinical applications: by relating the ANN’s algorithmic outcome to the user’s a-priori understanding of pathology-causality, they can strengthen confidence in the predictor, thereby increasing its clinical acceptance. To achieve this goal, additional restrictions and clarifications are crucial. Qualitatively, such maps need to be \emph{informative} for its users, i.e. narrow down regions of medical interest, hence coincide with medical knowledge and expectations \cite{lombrozo2006}. Furthermore, the regions' characteristic, i.e. the meaning of  \emph{maximal influence}, must be clearly conveyed. Quantitatively, such LE need to be \emph{faithful} to the underpinning predictor, i.e. dependent on architecture, parametrization, and preconditions \cite{adebayo2018}. 

The dominant class of methods follow a \emph{direct approach}. Utilizing an ANN's assumed analytic nature and its layered architecture, they typically employ a modified backpropagation approach to backtrack the ANN's activation to the input image \cite{simonyan2013,zhou2016}. While efficiently applicable, the resulting maps lack a clear a-priori interpretation, are potentially incomplete, coarse, and may deliver misleading information \cite{adebayo2018,dabkowski2017,fong2019,zintgraf2017}. Thereby they are potentially neither \emph{informative} nor \emph{faithful}, thus pose an inherent risk in medical environments. 

In contrast, \emph{reference based} LE approaches directly manipulate the input image and analyze the resulting prediction's differences \cite{fong2019}. They aim to assess an image-region's influence on prediction by counterfactual reasoning: how would the prediction score vary, if the region's image-information would be missing, i.e. its contribution marginalized? The prevailing heuristic approaches, e.g. Gaussian noise and blurring or replacement by a predefined colour \cite{zeiler2017,dabkowski2017,fong2019}, have been advanced to local neighborhood \cite{zintgraf2017} and stronger conditional generative models \cite{chang2018,uzunova2019}. Reference based LEs have the advantage of an a-priori clear and intuitively conveyable meaning of their result, hence address \emph{informativeness} for end-users. However, their applicability for medical imaging hinges on the utilized marginalization technique, i.e. the mapping between potentially pathological tissue representations and their healthy equivalent. Resulting \emph{prediction-neutral} regions need to depict healthy tissue per definition. Contradictory, the presented approaches introduce noise and thereby possibly pathological indications or anatomically implausible tissue (cf. Fig.~\ref{fig:inpainting_ov}). Hence, they violate the needed \emph{faithfulness} \cite{fong2019}. 

While dedicated generative adversarial networks (GANs) for medical images deliver significantly improved results, applications are hindered by possible resolutions and limited control over the globally acting models \cite{andermatt2019,baumgartner2017,becker2019,bermudez2018}. In \cite{major20}, the locally acting, but globally conditioned, per-pixel reconstruction of partial convolution inpainting (PCI) \cite{liu2018} is favoured over GANs, thereby enforcing anatomically sound, image specific replacements. While overcoming out-of-domain issues, this gradient descent based optimization method works iteratively, hence cannot be used in time restrictive environments.

\textbf{Contribution:} We introduce a \emph{resource efficient} reference based \emph{faithful} and \emph{informative} attribution method for real time pathology classifier interpretation. Utilizing a specialized ANN and exploiting PCI's local per-pixel reconstruction, conditioned on a global healthy tissue representation, we are able to enforce anatomically sound, image specific marginalization, without sacrificing computational efficiency. We formulate the ANN's objective function as a quantitative prediction problem under strict area constraints, thereby clarifying the resulting attribution map's a-priori meaning. 
We evaluate the approach on public mammography data and compare against two existing state-of-the-art methods. Furthermore, we exemplify the method's generalizability by demonstrating results on a second unrelated task, namely chest X-ray data. Our solution shows, both quantitatively and qualitatively, a significant reduction of localization ambiguity and clearer conveying results without sacrificing time efficiency.

\section{Methods}
Given a pathology classifier's prediction for an input image, we want to estimate its cause by attributing the specific pixel-regions that substantially influenced the predictor’s outcome. Informally, we search for the image-area that, if changed, results in a \emph{sufficiently healthy} image able to \emph{fool the classifier}. The resulting attribution-map needs to be \emph{informative} for the user and \emph{faithful} to its underpinning classifier. While we can quantitatively test for the latter, the former is an ill-posed problem. We therefore formalize as follows: 

Let $I$ denote an image of a domain $\mathcal{I}$ with pixels on a discrete grid $m_1 \times m_2$, $c$ a fixed pathology-class, and $f$ a classifier capable of estimating $p(c|I)$, the probability of $c$ for $I$. Also, let $M$ denote the attribution-map for image $I$ and class $c$, hence $M \in M^{m_1 \times m_2}(\{0,1\})$. Furthermore, assume a function $\pi(M)$ proficient  in marginalizing all pixel regions attributed by $M$ in $I$ such that the result of the operation is still within the domain of $f$. Hence, $\pi(M)$ yields a new image similar to $I$, but where we know all regions attributed by $M$ to be healthy per definition. Therefore, assuming $I$ depicts a pathological case and $M$ attributes only pathology pixel representations, $\pi(M)$ is a healthy counterfactual image to $I$. In any case $p(c|\pi(M))$ is well defined. Using this notation, we can formalize what an \emph{informative} map $\hat{M}$ means, hence give it an a-priori, testable semantic meaning. We define it as
\begin{equation*}
\hat{M}: = \argmin_{M \in \hat{\mathcal{M}}} d(M)\quad  \text{where}\quad  \hat{\mathcal{M}} := \{ p(c|\pi(M)) \leq \theta, d(M) \leq \delta, M \in \mathcal{S} \},
\end{equation*}
where $\theta$ is the classification-threshold, $d$ a metric measuring the attributed area, $\delta$ a constant limiting the attributed area, and $\mathcal{S}$ the set of compact and connected masks. Any map of $M^{m_1 \times m_2}(\{0,1\})$ can be (differentiably) mapped into $\mathcal{S}$ by taking the smoothed maximum of a convolution with a Gaussian kernel \cite{lange2014,fong2019}. In this form, $\hat{M}$ is clearly defined, and can be intuitively understood by end-users. 

Solving for $\hat{M}$ requires choosing (i) an appropriate measure $d$ (e.g. the map area in pixels), (ii) an appropriate size-limit $\delta$ (e.g. $n$ times average mass-size for mammography), and (iii) a fitting marginalization technique $\pi(\cdot)$. In the following we describe how we solve for $\hat{M}$ through an ANN, and overcome the out-of-domain obstacles by partial convolution \cite{liu2018} for marginalization.

\begin{figure}[t]
\begin{subfigure}[b]{1\textwidth}
\includegraphics[width=1\textwidth]{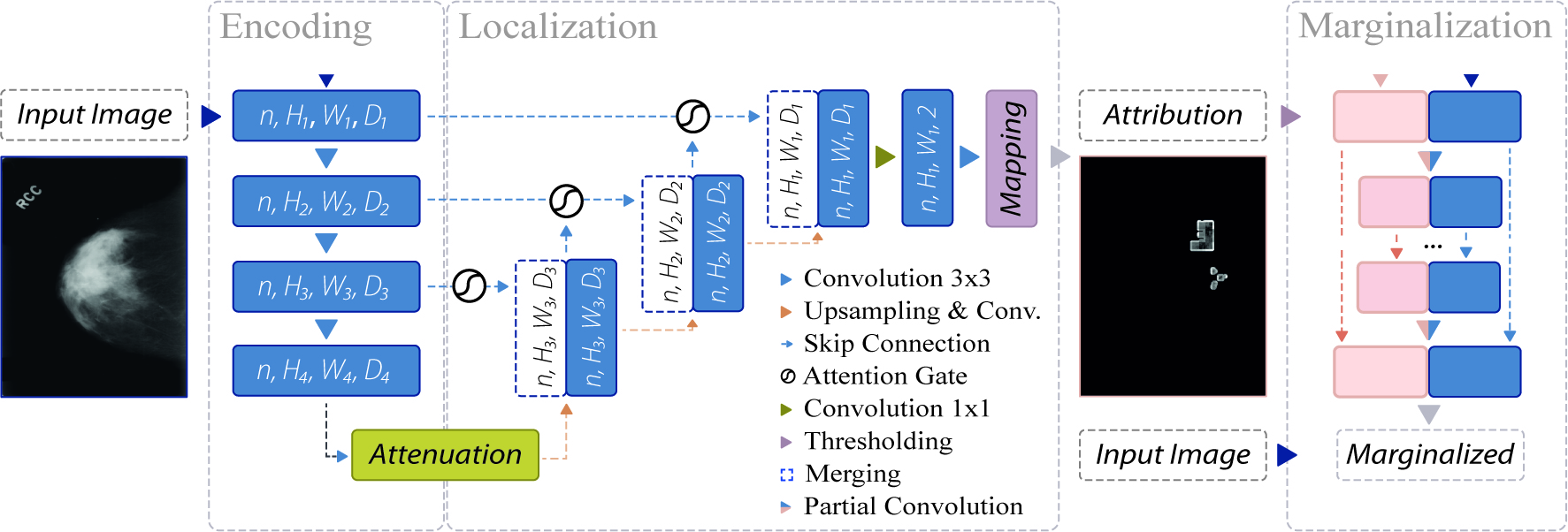}
\end{subfigure}

\caption{Attribution framework: The input image is encoded using the classifier’s features (left) and attenuated to enclose pathological regions (middle). During training, counterfactual images are produced by the marginalization-net (right), fed by  thresholded attribution (pink blocks) and  input image (blue blocks).}
\label{fig:architecture}
\end{figure}

\subsection{Architecture}

Iteratively finding solutions for $\hat{M}$ is typically time-consuming \cite{fong2019,major20}. Therefore, we develop a dedicated ANN, capable of finding the desired attribution in a single forward pass. To this end, the network learns on multiple resolutions, to combine relevant classifier-extracted features (cf. Fig.~\ref{fig:architecture}).
Inspired by \cite{dabkowski2017}, we build on a U-Net architecture, where the down-sampling, encoding branch consists of the trained classifier without its classification layers. These features, $x_{i,j,l}$, are subsequentially passed through a feature-filter, performing   $ x_{i,j,l} \cdot  \sigma ( (W_m \rho (W_{l}^\intercal  x_{i,j,l} + b_{l}) + b_{m} )) $ where $\rho$ is an element-wise nonlinearity (namely a rectified linear unit), $\sigma$ a normalization function (sigmoid function) and $W_.$ resp. $b_.$ linear transformation parameters. This is similar to additive attention, which, compared to multiplicative attention, has shown better performance on high dimensional input-features \cite{schlemper2019}.
The upsampling branch consists of four consecutive blocks of: upsampling by a factor of two, followed by convolution and merging with attention-gate weighted features from the classifier of the corresponding resolution scale.  After final upsampling back to input-resolution, we apply $1\times1$ conv. of depth two, resulting in two channels $c_{1,2}$. The final attribution-map $\hat{M}$ is derived through thresholding $ \frac{|c_1|}{|c_1| + |c_2|}$. Intuitively, the network attenuates the classifier's final features, generating an initial localization. This coarse map is subsequently refined by additional weighting and information from higher resolution features (cf. Fig.~\ref{fig:architecture}). We train the network, by minimizing
\begin{equation*}
 \mathcal{L}(M) = \phi(M) + \psi(M) + \lambda \cdot \mathcal{R}(M), \  \text{s.t.} \ d(M) \leq \delta
 \end{equation*}
 where $\phi(M) := -1 \cdot \log ( p(c| \pi(M)) )$, $ \psi(M) := \log (\text{odds}(I)) - \log (\text{odds}(\pi(M))) $, and $\text{odds}(I) = \frac{ p(c|I) }{ 1 - p(c|I)}$, hence weigh the probability of the marginalized image, enforcing $ p(c|\pi(M)) \leq \theta$. We introduced an additional regularization-term: a weighted version of total variation \cite{peng2019}, which experimentally greatly improved convergence. All terms where normalized through a generalized logistic function. The inequality constraint was enforced by the method proposed in \cite{kervadec2018}. Note that after mapping into $\mathcal{S}$, any solution to $\mathcal{L}$ will also estimate $\hat{M}$, thereby yielding our desired attribution-map. The parametrization is task/classifier-dependent and will be described in the following sections.

\subsection{Marginalization}
As we need to derive $p(c| \pi(M))$, our goal is to marginalize arbitrary image regions marked by our network during its training process. Therefore, we aim for an image inpainting method to replace pathological tissue by healthy appearance. The result should resemble valid global anatomical appearance with high quality local texture.
To address the these  criteria we apply the U-Net like architecture with partial convolution blocks of \cite{liu2018} which gets an image and a hole mask as input (cf. Fig.~\ref{fig:architecture}). Partial convolution considers only unmasked inputs in a current sliding window to compute its output. Where it succeeded, hole mask positions are eliminated. This mechanism helps conditioning on local texture. The loss function ($\mathcal{L}_{PCI}$) balances local per-pixel reconstruction quality of masked/unmasked regions ($ \mathcal{L}_{hole} /\mathcal{L}_{valid}  $), against globally sound anatomical appearance ($\mathcal{L}_{perc}, \mathcal{L}_{style} $). An additional total variation term ($\mathcal{L}_{tv}$) ensures a smooth transition between hole and present image regions in the final result. This yields  $\mathcal{L}_{PCI} = \mathcal{L}_{valid} + 6\cdot\mathcal{L}_{hole} + 0.05\cdot \mathcal{L}_{perc} + 120\cdot \mathcal{L}_{style} + 0.1\cdot \mathcal{L}_{tv}$ where parametrization follows~\cite{liu2018}.
The architecture's contraction path consists of 8 partial convolution blocks with a stride of 2. The kernels of depth 64, 128, 256, 512, $\ldots$, 512 have sizes 7, 5, 5, 3, $\ldots$, 3. The expansion path, a mirrored version of the contraction path, contains upsampling layers with a factor of 2, kernel size of 3 at every layer, and a final filterdepth of 3. 
Each block contains batch normalization (BN) and ReLU/LeakyReLU (alpha=0.2) activations in the contraction/expansion paths which are connected by skip connections.
Zero padding of the input was applied to control resolution shrinkage and keep aspect ratio.

\section{Experimental Setup}

\textbf{Datasets:} We evaluated our framework on two different datasets, on mammography scans and on chest X-ray images. For mammography, we complemented the 1565 annotated, pathological CBIS-DDSM scans containing masses~\cite{lee2016} with 2778 healthy DDSM images~\cite{heath2000} and downsampled them to 576x448 pixels.
Data was split into 1231/2000 mass/healthy samples for training, and into 334/778 scans for testing. 
There was no patient-wise overlap between the training/test data. %
We demonstrate generalization on a private collection of healthy and tuberculotic (TBC) frontal chest X-ray images, at a downsampled resolution of 256x256. We split healthy images into sets of 1700/135 for training respectively validation set, and TBC cases into 700/70. The test set contains 52 healthy and 52 TBC samples. No pixel-wise GT information was provided for this data.

\textbf{Classifiers:} The backbone of our mammography attribution network is a MobileNet~\cite{howard2017} classifier for distinguishing between healthy samples and scans with masses. The network was trained
using the Adam optimizer with batchsize of 4 and learning rate of 1e-5 for 250 epochs with early stopping. The network was pretrained with 50k 224x224 pixel patches from the training data for the same task.
The TBC attribution utilized a DenseNet-121~\cite{huang2017} classifier for the binary classification task of healthy or TBC cases. It was trained using the SGD momentum optimizer with a batchsize of 32 and learning rate of 1e-5 for 2000 epochs.
This network was pretrained on the CheXpert dataset~\cite{irvin2019}.

\textbf{Marginalization:} The chest X-ray images have one magnitude smaller resolution than the mammography scans, thus we removed the bottom-most blocks from the contraction and expansion paths. Both inpainter networks were trained on healthy training samples with a batch size of 1 for mammography and 5 for chest X-ray. Training was done in two phases, the first phase with BN after each partial convolution layer and the second with BN only in the expansion path. The network for the mass classification task was trained with learning rates of $1\text{e-}5$/$1\text{e-}6$ and for the TBC classification task of $2\text{e-}4$/$1\text{e-}5$ for the two phases. For each image irregular masks were generated which mimic possible configurations during the attribution network training~\cite{liu2018}.

\textbf{Attribution:} We used the last four resolution-scales of each classifier, and in all cases the features immediately after the activation function, following the convolution. The weights of the pre-trained ANNs were kept fixed during the complete process. Filterdepths of the upsampling convolution blocks correspond to the equivalent down-sampling filters, filter-size is fixed to $1\times1$. Upsampling itself is done via neighborhood upsampling. We used standard gradient descent, and a cyclic learning rate~\cite{smith2015}, varying between $1\text{e-}6$ and $1\text{e-}4$, and trained for up to $5000$ epochs with early stopping. We thresholded the masks at $0.55$, and used a Gaussian RBF with $\sigma=5\text{e-}2$, and a smoothing parameter of $30$. All trainable weights where random-normal initialized.

\section{Results and Conclusion}
\textbf{Marginalization:} To evaluate the inpainter network we assessed how much the classification score of an image changes, when pathological tissue is replaced. 

Thus, we computed ROC curves using the classifier on all test samples (i) without any inpainting as reference, and for comparison, randomly sampled inpainting  (ii) only in healthy respective (iii) pathological scans over 10 runs (Fig.~\ref{fig:inpainting_ov}).
The clear distance between the ROC curves of the mammography image classifiers without any inpainting, yielding an AUC of 0.89, and with inpainting in pathological regions, resulting in an AUC of 0.86, shows that the classifier is sensitive to changes around pathological regions of the image.
Moreover, it is visible that the ROC curves of inpainting in healthy tissues with an AUC of 0.89 follow closely the unaffected classifier's ROC curve (Fig.~\ref{fig:inpainting_ov}).
The AUC scores for the TBC classifier without and with inpainting in healthy tissue are 0.89 and 0.88 which proves the above mentioned observations. Pathological tissue inpainting was ommitted in this case due to the lack of pixel-wise annotations.

\begin{figure*}[htb!]
\centering
  \begin{subfigure}[b]{.23\linewidth}
    \centering
    \includegraphics[width=.99\textwidth,trim={0 40 0 50},clip]{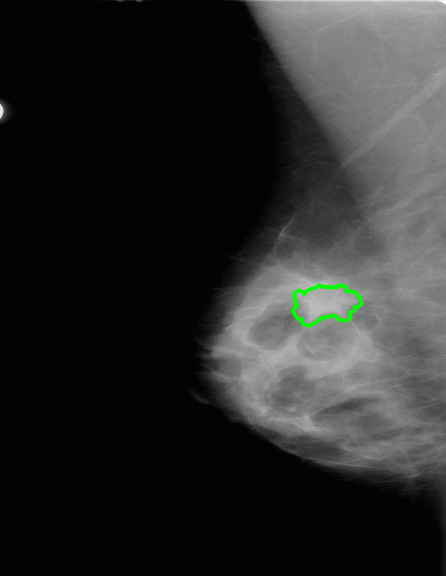}
  \end{subfigure}%
  \begin{subfigure}[b]{.23\linewidth}
    \centering
    \includegraphics[width=.99\textwidth,trim={0 40 0 50},clip]{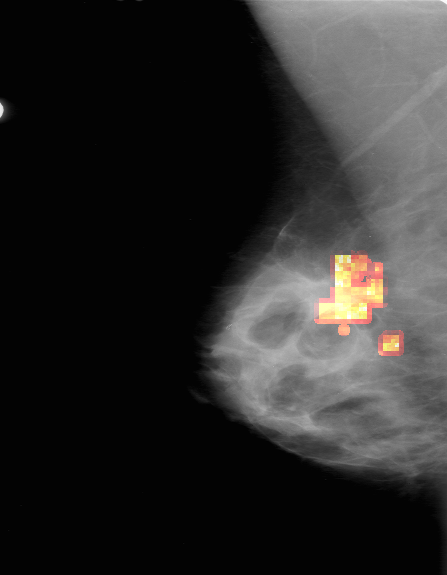}
  \end{subfigure}%
  \begin{subfigure}[b]{.23\linewidth}
    \centering
    \includegraphics[width=.99\textwidth,trim={0 40 0 50},clip]{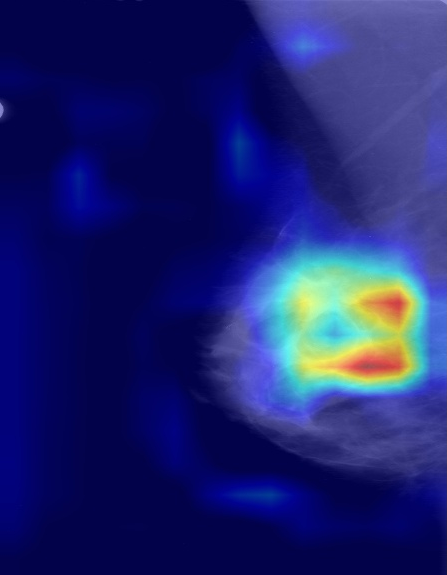}
  \end{subfigure}%
	\begin{subfigure}[b]{.23\linewidth}
    \centering
    \includegraphics[width=.99\textwidth,trim={0 40 0 50},clip]{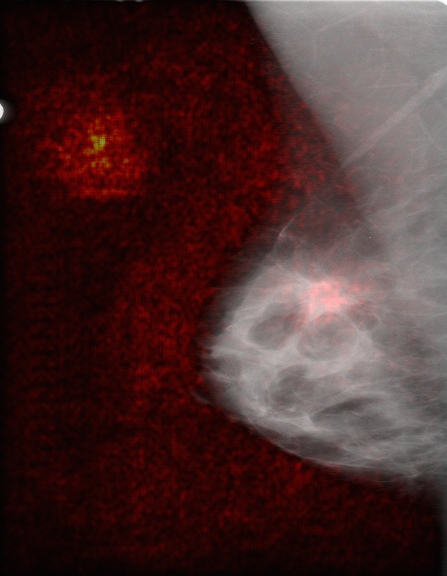}
  \end{subfigure}%
	
	\begin{subfigure}[b]{.23\linewidth}
    \centering
    \includegraphics[width=.99\textwidth,trim={0 30 0 40},clip]{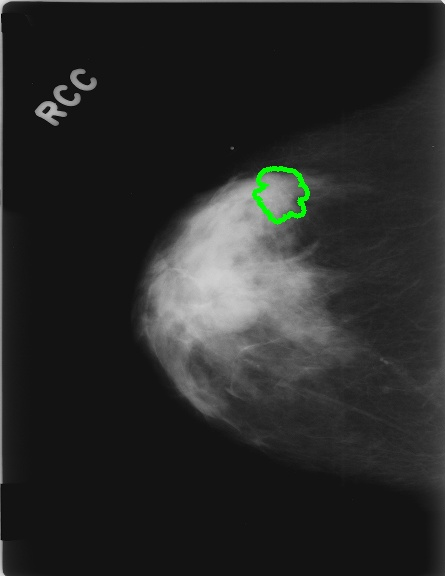}
  \end{subfigure}%
  \begin{subfigure}[b]{.23\linewidth}
    \centering
    \includegraphics[width=.99\textwidth,trim={0 30 0 40},clip]{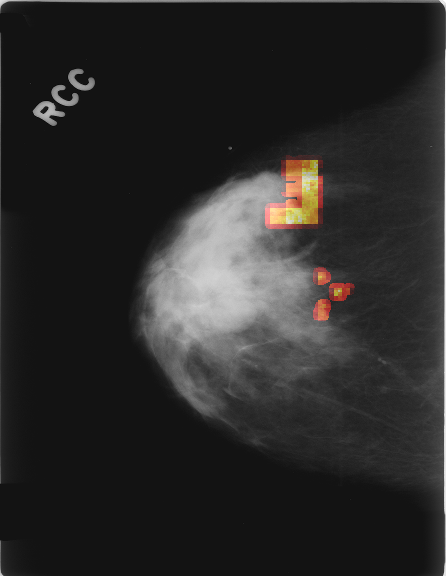}
  \end{subfigure}%
  \begin{subfigure}[b]{.23\linewidth}
    \centering
    \includegraphics[width=.99\textwidth,trim={0 30 0 40},clip]{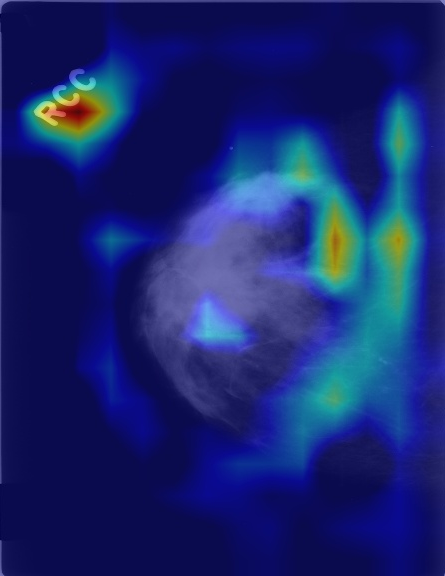}
  \end{subfigure}%
	\begin{subfigure}[b]{.23\linewidth}
    \centering
    \includegraphics[width=.99\textwidth,trim={0 30 0 40},clip]{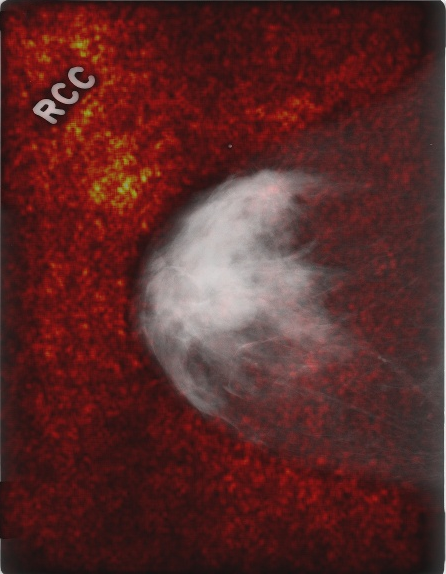}
  \end{subfigure}%
	
	\begin{subfigure}[b]{.23\linewidth}
    \centering
    \includegraphics[width=.985\textwidth]{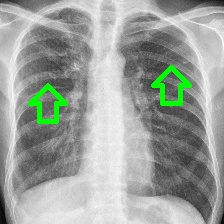}
    \caption{}\label{fig:1e}
  \end{subfigure}%
  \begin{subfigure}[b]{.23\linewidth}
    \centering
    \includegraphics[width=.985\textwidth]{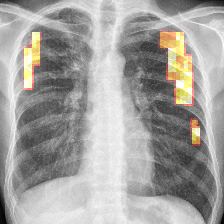}
    \caption{}\label{fig:1f}
  \end{subfigure}%
  \begin{subfigure}[b]{.23\linewidth}
    \centering
    \includegraphics[width=.985\textwidth]{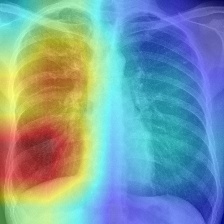}
    \caption{}\label{fig:1g}
  \end{subfigure}%
	\begin{subfigure}[b]{.23\linewidth}
    \centering
    \includegraphics[width=.985\textwidth]{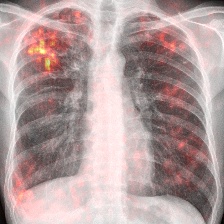}
    \caption{}\label{fig:1h}
  \end{subfigure}%

\caption{Result attribution heatmaps for mammography~\cite{lee2016} and chest X-ray~\cite{jaeger2014}: (a) original image overlayed with annotation contours (and arrows for missing GT), (b) our attribution framework. (c) GradCAM \cite{selvaraju2017} (d) Saliency \cite{simonyan2013}.} 
\label{fig:results}
\end{figure*}

\textbf{Attribution:} We compared our attribution network against the gradient explanation \emph{saliency map} \cite{simonyan2013} (SAL), and the network/gradient-derived \emph{GradCAM} \cite{selvaraju2017} visualizations. We limited our comparisons to these direct approaches, as they are widely used within medical imaging \cite{irvin2019}, and inherently valid \cite{adebayo2018}. Popular \emph{reference based} approaches either utilize blurring, noise or some other heuristic \cite{fong2019,dabkowski2017,zintgraf2017}, or were not available \cite{chang2018}, therefore could not be considered. Quantitatively, we relate (i) the result-maps $\hat{M}$ to both organ, and ground truth (GT) annotations, and (ii) to each other. Particularly for (i) we studied the Hausdorff distances $H$ between GT and $\hat{M}$ indicating location proximity. Lower values demonstrate better localization in respect to the pathology. Further, we performed a weak localization experiment \cite{dabkowski2017,fong2019}: per image, we derived bounding boxes (BB) for each connected component of GT and $\hat{M}$ attributions. A GT BB counts as found, if any $\hat{M}$ BB has an \emph{IOU} $\leq0.125$. We chose this threshold, as a proficient classifier presumably focuses on the masses' boundaries and neighborhoods, thereby limiting possible BB-overlap. We report average localization $L$. For (ii) we derived the area ratio $A$ between $\hat{M}$ and organ-mask (breast-area) or whole image (chest X-ray). Again, lower values indicate a smaller thereby clearer map. Due to missing GT we could only derive (ii) for TBC. All measurements were performed on binary masks, hence GradCAM and SAL had to be thresholded. We chose the $50, 75, 90$ percentiles, i.e. compared $50, 25, 10$ percent of the map-points. Where multiple pathologies, or mapping results occurred we used the median for a robust estimation per image. Statistically significant difference between all resulting findings was formalized using Wilcoxon signed-rank tests, for $ \alpha < 0.05$. Additionally we followed \cite{adebayo2018}, and tested our network with randomised parametrization (labels have no effect in our case).

\begin{table}[tp]
\centering
\setlength{\tabcolsep}{0.5em}

\begin{tabular}{|c|c|c|c||c|c|c|}
\hline

P & \textit{$H_{ours}$} & \textit{$H_{grad}$} & \textit{$H_{sal}$}  & \textit{$L_{ours}$} & \textit{$L_{grad}$} & \textit{$L_{sal}$} \\
\hline
50  &\textbf{188.12}$\pm$68.3 & 296.29$\pm$54.4& 240.83$\pm$36.2 & \textbf{0.45}  &  0.06 & 0.27  \\
\hline
75 &\textbf{188.12}$\pm$68.3 & 274.86$\pm$40.0& 257.85$\pm$38.6 &  \textbf{0.45}  &  0.23 &  0.30 \\
\hline
90 &\textbf{188.12}$\pm$68.3 & 243.80$\pm$59.6& 259.57$\pm$43.7 & \textbf{0.45}  & 0.28   &  0.25 \\
\hline
\end{tabular}

\begin{tabular}{|c|c|c|c||c|c|c|}
\hline

P & \textit{$A_{ours}^{mammo}$} & \textit{$A_{grad}^{mammo}$} & \textit{$A_{sal}^{mammo}$}  & \textit{$A_{ours}^{tbc}$} & \textit{$A_{grad}^{tbc}$} & \textit{$A_{sal}^{tbc}$} \\
\hline
50  & \textbf{0.07}$\pm$0.04  &  1.10$\pm$0.10 & 1.10$\pm$.14  &\textbf{0.06}$\pm$0.0 & 0.50$\pm$0.0 & 0.50$\pm$0.0  \\
\hline
75 & \textbf{0.07}$\pm$0.04  &  0.55$\pm$0.21 &  0.55$\pm$0.2 & \textbf{0.06}$\pm$0.0 & 0.25$\pm$0.0 & 0.25$\pm$0.0 \\
\hline
90 & \textbf{0.07}$\pm$0.04 & 0.22$\pm$0.40  &  0.22$\pm$0.43  &\textbf{0.06}$\pm$0.0 & 0.10$\pm$0.0 & 0.10$\pm$0.0 \\
\hline
\end{tabular}


\caption{Top: Hausdorff distances $H$ and weak localization results $L$, relating maps $\hat{M}$ to GT ; Bottom: relating maps $\hat{M}$ to the organ resp. image-size  } 
\label{tab:results_mammo}
\end{table}

As seen in Table~\ref{tab:results_mammo}, our framework achieves significantly lower $H$, than either GradCAM or SAL at all threshold levels. Moreover, we report significantly better weak localization ($L$) which underlines the higher accuracy of our approach. Qualitatively our attribution-maps are tighter focused (c.f. Fig.~\ref{fig:results}(b)) and enclose the masses. The former is also expressed by the lower overlap values $A$.  All p-values where significantly below $1\text{e-}2$, hardening our results. Randomization of the ANN's weights yields pure noise maps, hence we pass \cite{adebayo2018}'s checks.

\textbf{Timing:} We estimated the time needed for a single attribution map, one forward pass, by averaging over ten times repeated map derivations for all images of the resp. test sets. These were compared with the analogous timings of GRAD and SAL. Additionally, as a reference for iterative methods, we compared with \cite{major20} that, using same marginalization technique, yields equivalent maps. 

Our model is capable of deriving $75$ mammography maps per second (mps) utilizing a GPU (NVIDIA Titan RTX). This compares favourably to both GRAD and SAL, 50 resp. 31 mps, and significantly outperforms the iterative method (27 seconds per map). Considering the smaller X-ray images, these throughputs increase up to a factor of three, sufficient even for  real time environments.

\textbf{Conclusion:} In this work, we proposed a novel neural network based attribution method for real time interpretation of pathology classifiers. Our reference based approach enforces domain aware marginalization, without sacrificing computational efficiency. Overcoming these common obstacles, our approach can provide further confidence, and thereby increase critical user acceptance. We compared our method with state-of-the-art techniques on two different tasks, and show  favorable results throughout. This underlines the suitability of our approach as an interpretation tool in radiology workflows.

%
%
%
%
\bibliographystyle{splncs04}
\bibliography{visNet}

\end{document}